%% file: arxiv.tex
\documentclass{article}

\usepackage{arxiv}

\usepackage[utf8]{inputenc} 
\usepackage[T1]{fontenc}    
\usepackage{hyperref}       
\usepackage{url}            
\usepackage{booktabs}       
\usepackage{amsfonts}       
\usepackage{nicefrac}       
\usepackage{microtype}      
\usepackage{lipsum}

\usepackage{graphicx}
\graphicspath{{figures/}}
\usepackage{multirow}
\usepackage{algorithm}
\usepackage{algorithmic}
\usepackage{setspace}
\usepackage{subfigure}
\usepackage{optidef}    
\usepackage{siunitx}    
\usepackage{interval}   
\usepackage{upgreek}    
\usepackage[table,x11names]{xcolor} 
\usepackage{wrapfig}
\usepackage{changes}
\usepackage{sidecap}
\usepackage{tikz}
\usetikzlibrary{calc}

\definechangesauthor[name={Reviewer 1. Comment 1}, color=red]{r11}
\definechangesauthor[name={Reviewer 1. Comment 2}, color=red]{r12}
\definechangesauthor[name={Reviewer 1. Comment 4}, color=red]{r14}
\definechangesauthor[name={Reviewer 1. Comment 4}, color=red]{r16}
\definechangesauthor[name={Reviewer 2.}, color=blue]{r2}
\definechangesauthor[name={Reviewer 4. Comment 4}, color=black!50!green]{r44}

\newcommand{\finalstate}[1][]{%
  \renewcommand{\added}[2][]{##2}%
  \renewcommand{\replaced}[3][]{##2}}%
\finalstate 

\def\centerarc[#1](#2)(#3:#4:#5)
    { \draw[#1] ($(#2)+({#5*cos(#3)},{#5*sin(#3)})$) arc (#3:#4:#5); }

\title{Learning the optimal state-feedback via supervised imitation learning}

\author{
  Dharmesh Tailor, Dario Izzo\thanks{Dario Izzo is the corresponding author} \\
  Advanced Concepts Team \\
  European Space Agency \\
  Noordwijk, 2201 AZ \\
  Netherlands \\
  \texttt{dharmesh.tailor@live.co.uk} \\
  \texttt{dario.izzo@esa.int} \\
}

\begin{document}
\maketitle

\begin{abstract}
\input{00_abstract}
\end{abstract}


\section{Introduction}
\input{01_introduction.tex}

\section{Quadcopter model}
\label{sec:quadcopter}
\input{02_quadcopter.tex}

\section{State-feedback approximation}
\input{03_policy.tex}

\section{Deep neural network training}
\label{sec:dnn}
\input{04_dnn.tex}

\section{Training experiments}
\input{05_experiments.tex}

\section{Policy evaluation}
\input{06_policy_eval.tex}

\section{Softplus units}
\input{07_softplus.tex}

\section{Discussion and future work}
\input{08_conclusion.tex}

\bibliographystyle{acm}  
\bibliography{refs}  

%
%
%
%

\end{document}

%% file: 00_abstract.tex
Imitation learning is a control design paradigm that seeks to learn a control policy reproducing demonstrations from expert agents.
By substituting expert demonstrations for optimal behaviours, the same paradigm leads to the design of control policies closely approximating the optimal state-feedback.
This approach requires training a machine learning algorithm (in our case deep neural networks) directly on state-control pairs originating from optimal trajectories.
We have shown in previous work that, when restricted to low-dimensional state and control spaces, this approach is very successful in several deterministic, non-linear problems in continuous-time.
In this work, we refine our previous studies using as a test case a simple quadcopter model with quadratic and time-optimal objective functions. We describe in detail the best learning pipeline we have developed, that is able to approximate via deep neural networks the state-feedback map to a very high accuracy. We introduce the use of the softplus activation function in the hidden units of neural networks showing that it results in a smoother control profile whilst retaining the benefits of \replaced[id=r2]{rectifiers}{ReLUs}.
We show how to evaluate the optimality of the trained state-feedback, and find that already with two layers the objective function reached and its optimal value differ by less than one percent. We later consider also an additional metric linked to the system asymptotic behaviour - time taken to converge to the policy's fixed point.
With respect to these metrics, we show that improvements in the mean absolute error do not necessarily correspond to better policies.

%% file: 01_introduction.tex
The dynamic programming approach to deterministic optimal control indicates the existence of an optimal state-feedback map or \textit{policy} \cite{kirk1970optimal}.
This is a consequence of the Hamilton-Jacobi-Bellman (HJB) equation in continuous-time settings.
If the solution is pursued in the viscosity sense \cite{bardi2008optimal} then such a solution is also unique, making the optimal control problem (in terms of the value function) well-posed in the Hadamard sense \cite{hadamard1902problemes}.
However, for most problems of interest analytic solutions to the HJB equation cannot be found and approximate methods are resorted to (e.g. \cite{beard1997galerkin}).
This is particularly the case for systems modelled by non-linear dynamics or where the cost function is not quadratic in the state and control.

The alternate approach to optimal control is Pontryagin's minimum principle \cite{pontryagin2018mathematical} in which the optimal solution sought is one defined between two states (or sets) only (i.e. a single trajectory).
When applicable, one can use the solutions coming from the repeated use of Pontryagin's principle to learn an approximation to the solution of the HJB equations \cite{sanchez2018real}.
\added[id=r12]{%
This could be viewed as a form of imitation learning (see \cite{pomerleau1989alvinn} for an example of early pioneering work) defined as \textit{"[training] a classifier or regressor to replicate an expert's policy given training data of the encountered observations and actions performed by the expert"} \cite{ross2010efficient}.
In our case the encountered observations and actions are substituted for optimal trajectories.}
In general, learning a policy from optimal control actions is a troublesome approach as reported in \cite{mordatch2014combining}. This has been attributed to the fact that the distribution of states the policy encounters during execution differs from the distribution the policy is trained on \cite{ross2011reduction}. To overcome the issue, various methods are proposed e.g. DAGGER \cite{ross2011reduction}.
These methods share in common the ability for the policy being learned to influence the trajectories (and therefore states) contained within the training dataset. This is achieved by iterating between trajectory optimisation and policy execution whereby the states encountered constrain the optimal trajectories.
In more recent years, the use of a trajectory optimiser to aid policy learning has been investigated within reinforcement learning e.g. \cite{levine2013guided, levine2013variational} with great success.
It was reported that even DAGGER-like approaches are unsuccessful at tasks involving contact dynamics or with high-dimensional state/action spaces.
Instead it was proposed to augment the objective function of the trajectory optimiser with a term measuring the deviation of the controls from the current policy.
Such work was intended to fix the issues that arise when learning a policy from a database of optimal trajectories. S\'anchez-S\'anchez, et al. (2018) \cite{sanchez2018real} demonstrated that when applied to low-dimensional problems and, in particular, systems that converge to a small region of the state space, such issues do not arise and a straightforward approach is therefore desirable and very successful. This is attributed to the scaling up of the number of trajectories used resulting in a dense coverage of the state space for relatively low dimensional problems. Intuitively this can be understood by considering a non-parametric policy in which control selection is based on a nearest-neighbour look-up in the library of trajectories \cite{stolle2006policies}.

Effective learning from such large datasets was achieved using deep neural networks trained by stochastic gradient descent. It is worth noting here that feedforward networks were used despite training for a task resembling sequence prediction. This choice is justified by the Bellman optimality principle and can also be seen by considering that the solution to the HJB equation depends solely with respect to the current state.
Despite this, neural networks with a recurrent architecture have been used in a similar context, for example in \cite{furfaro2018recurrent}), and could be advantageous when modelling errors or imperfect sensing are considered during simulation. 

\added[id=r11]{%
In this work we build upon the work of S\'anchez-S\'anchez, et al. (2018) \cite{sanchez2018real} using as a test case the two-dimensional quadcopter model with quadratic and time-optimal objective functions considered there.
This model was chosen due to its simplicity, to avoid this work becoming too involved in the intricacies of a particular control problem.
The generality of the methods presented means the insights drawn are equally applicable to other control problems previously considered such as spacecraft pinpoint landing \cite{sanchez2018real} and orbital transfers \cite{izzo2018machine}.}
We describe in detail an improved learning pipeline and confirm that it is able to approximate the state-feedback map to a very high accuracy.
In order to determine the limits on this accuracy, we perform a wider search on the hyperparameters of the neural networks, with particular attention to the depth and number of units per hidden layer.
\added[id=r2]{%
We introduce the use of the softplus activation function in the hidden units showing how this results in a smoother and more meaningful control profile whilst retaining the benefits of rectifiers during training (e.g. improved gradient propagation in multi-layer networks).}
We propose a new method to evaluate the optimality of the trained state-feedback, avoiding the complexities that were introduced in \cite{sanchez2018real}, and find that already with two layers the objective function reached and its optimal value differ by less than one percent. 
With respect to this metric, we find that improvements in the mean absolute error do not necessarily correspond to improved policies. 
For many of the trained policies, we observe that the resulting system stabilises at a nearby equilibrium state after reaching the target state. 
This property forms the basis of a second metric we consider and that describes the asymptotic behaviour of the system, namely the time taken to converge to the stability point.

%% file: 02_quadcopter.tex
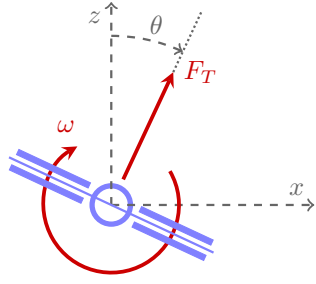
\begin{figure}[tb]
\centering
\begin{tikzpicture}[scale=1.5]
	\pgfmathsetmacro\th{-25}
	\pgfmathsetmacro\L{2}
    \pgfmathsetmacro\ycom{0.0}
    \pgfmathsetmacro\zcom{0.0}
    \pgfmathsetmacro\eps{0.05}
    \pgfmathsetmacro\rad{0.5}
    
	\draw[line width=0.7mm, blue!50] (\ycom, \zcom) circle (0.17);
	
    \draw[->, dashed, line width=0.3mm, black!60] (0,0) -- (1.8,0) node[above left] {$x$};
    \draw[->, dashed, line width=0.3mm, black!60] (0,0) -- (0,1.8) node[below left] {$z$};
    
    \centerarc[->, >=stealth, line width=0.5mm, black!20!red](\ycom,\zcom)(30:-240:0.6);
    \node[black!20!red] at (-0.4,0.7) {$\omega$};
	
	\begin{scope}[rotate around={\th:(\ycom,\zcom)}]
        \draw[-, line width=0.3mm, blue!50] (\ycom-\L/2,\zcom) -- (\ycom+\L/2,\zcom);
        \draw [fill=gray,blue!50] (\ycom-\L/2+\eps,\zcom+\eps) rectangle (\ycom-5*\eps,\zcom+2*\eps);
        \draw [fill=gray,blue!50] (\ycom-\L/2+\eps,\zcom-\eps) rectangle (\ycom-5*\eps,\zcom-2*\eps);
        \draw [fill=gray,blue!50] (\ycom+\L/2-\eps,\zcom+\eps) rectangle (\ycom+5*\eps,\zcom+2*\eps);
        \draw [fill=gray,blue!50] (\ycom+\L/2-\eps,\zcom-\eps) rectangle (\ycom+5*\eps,\zcom-2*\eps);
        \draw[->, >=stealth, line width=0.5mm, black!20!red] (\ycom, \zcom+5*\eps) -- (\ycom, \zcom + 1.3) node[right] {$F_T$};
        \draw[densely dotted, line width=0.3mm, black!60] (\ycom, \zcom + 1.3) -- (\ycom, \zcom + 1.9);
    \end{scope}
    
    \centerarc[->, dashed, line width=0.3mm, black!60](\ycom,\zcom)(90:90+\th:1.5);
    \node[black!60] at (0.4,1.6) {$\theta$};
\end{tikzpicture}
\caption{Two-dimensional model of a quadcopter considered in this study. Control inputs $\{F_T, \omega\}$ as well as the state variables $\{x, z, \theta\}$ are indicated.}
\label{fig:quad}
\end{figure}

We consider a two-dimensional model of a quadcopter with 3 degrees of freedom \cite{hehn2012performance}
in which the task is to manoeuvre the vehicle to a goal location (see Fig. \ref{fig:quad}).
The system states $\mathbf{x} \in \rm I\!R^{n_x}$ and control inputs $\mathbf{u} \in \rm I\!R^{n_u}$ ($n_x = 5, n_u = 2$) are defined to be:
\begin{equation}
    \mathbf{x}=[x, v_x, z, v_z, \theta]^{T}, \quad \mathbf{u} = [F_T, \omega]^{T}
\end{equation}
where $x$ is the horizontal position, 
$z$ is the vertical position,
$(v_x,v_z)$ is the velocity,
$\theta$ is the pitch angle,
$F_T$ is the total thrust force
and $\omega$ is the pitch rate.

The dynamics of the system is given by:
\begin{equation}
    \mathbf{\dot{x}} = 
    \begin{bmatrix}
        \dot{x} \\
        \dot{v}_x \\
        \dot{z} \\
        \dot{v}_z \\
        \dot{\theta}
    \end{bmatrix} =
    \mathbf{f}(\mathbf{x}, \mathbf{u}) =
    \begin{bmatrix}
        v_x \\
        \frac{F_T}{m} \sin{\theta} - \beta{v_x} \\
        v_z \\
        \frac{F_T}{m} \cos{\theta} - g - \beta{v_z} \\
        \omega
    \end{bmatrix}
\end{equation}
where $g$ is the gravitational acceleration and
$m$ is the mass of the quadcopter.
Compared with previous work~\cite{sanchez2018real}, the dynamics have been modified to take into account the effect of drag forces via a drag coefficient $\beta$.
The control inputs are constrained as follows:
\begin{equation}
    0 \leq F_T \leq F_T^{max}, \quad | \omega | \leq \omega^{max}
\end{equation}
where $F_T^{max}$ and $\omega^{max}$ are limits on the maximum allowable 
thrust and pitch rate respectively.

As per the standard formulation of the optimal control problem, 
the task of manoeuvring the quadcopter 
is equivalent to finding the state-control trajectory  $\{ \mathbf{x}(t), \mathbf{u}(t) : 0 \leq t \leq T \}$ satisfying:
\begin{mini}|l|
    {\mathbf{u(t)}, T}{J(\mathbf{x}(t), \mathbf{u}(t), T)}{}{}
    \addConstraint{\mathbf{\dot{x}}(t) = \mathbf{f}(\mathbf{x}(t), \mathbf{u}(t)) \quad \forall t}
    \addConstraint{\mathbf{x}(0) = \mathbf{x}_o}
    \addConstraint{\mathbf{x}(T) = \mathbf{x}_f}
\end{mini}
where $\mathbf{x}_o$ is the initial state, $\mathbf{x}_f$ is the target state,
$J$ is the objective function determining the path cost
and $T$ is the total trajectory time.
We consider two objective functions, the first is quadratic control,
\begin{equation}
\begin{aligned}
    J &= \int_0^{T} {|| \mathbf{u}(t) ||}^2 \, \text{d}t \\
      &= \int_0^{T} (F_T(t)^2 + \omega(t)^2) \, \text{d}t,
\end{aligned}
\end{equation}
and the second seeks to minimise time,
\begin{equation}
    J = T.
\end{equation}

There are multiple ways of solving optimal control problems of this form \cite{betts1998survey}.
We use a direct transcription and collocation method, 
namely Hermite-Simpson transcription,
which transforms the trajectory optimisation problem into a 
non-linear programming (NLP) problem \cite{betts2010practical}.
The modelling language AMPL was used to specify the NLP problem.
This is then solved using an NLP solver, many of which are supported by AMPL.
For our problem we used SNOPT \cite{gill2005snopt}, a sequental quadratic programming NLP solver.
We observe that this solver is able to converge to the optimal solution
starting from an arbitrary initial guess for most choices of the initial and target state.

\begin{figure}[!htbp]
	\centering
    \subfigure[]{
    	\label{subfig:control_profiles_state}
    	\includegraphics[width=0.48\textwidth]{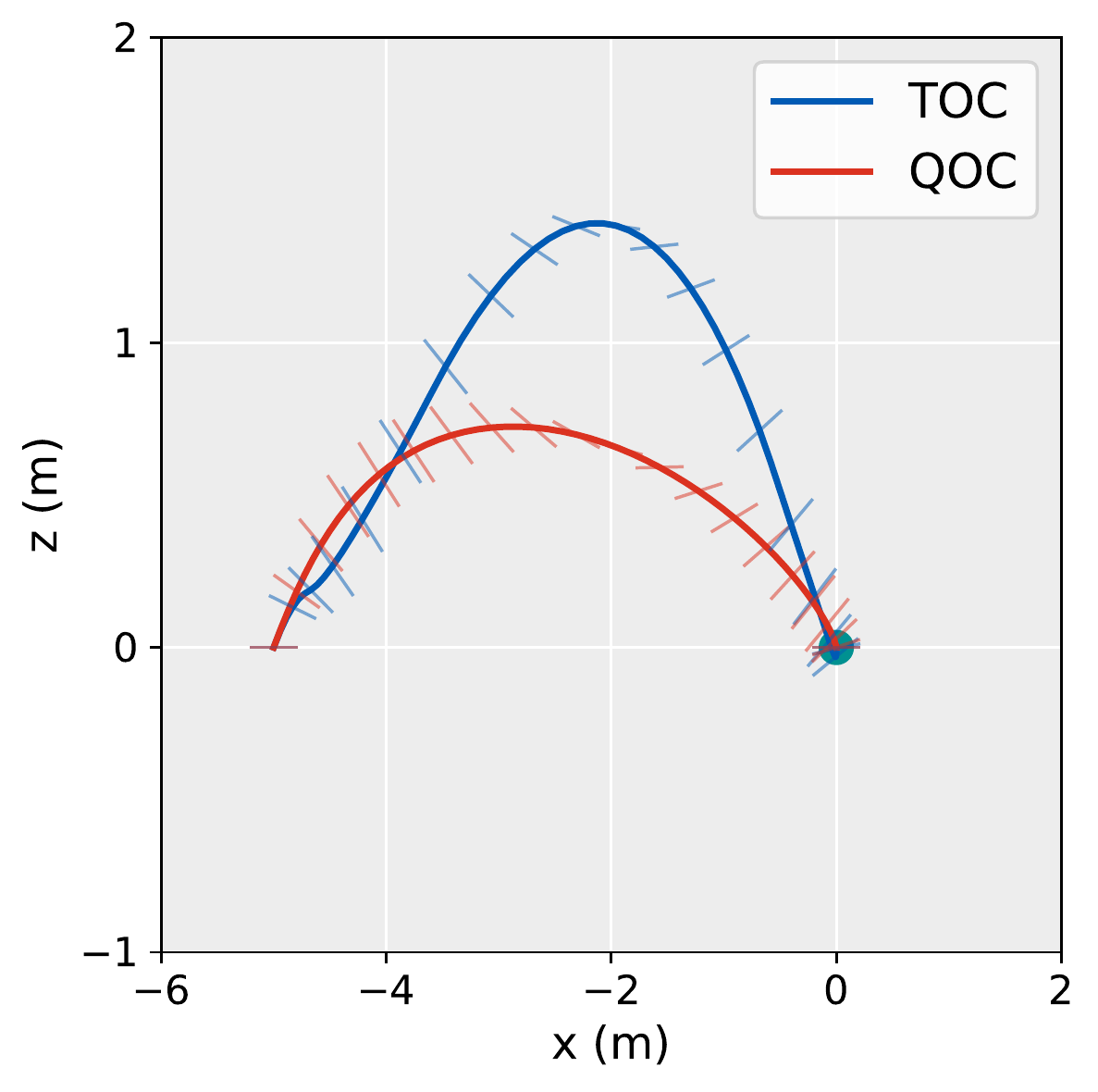}
    }
    \hfill
    \subfigure[]{
    	\label{subfig:control_profiles_control}
    	\includegraphics[width=0.48\textwidth]{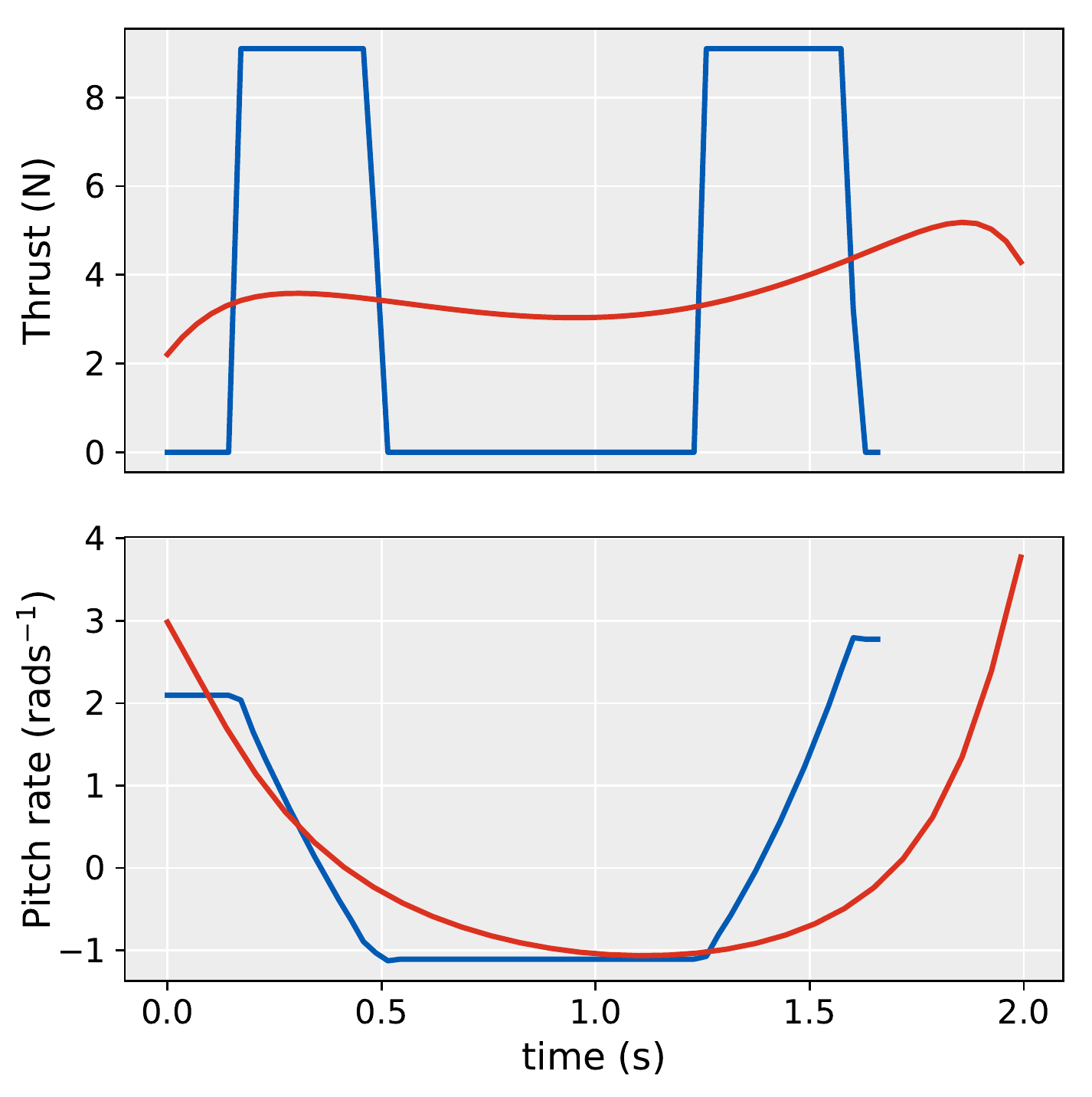}
    }
	\caption{
		Example state-control trajectory for the quadcopter model with quadratic (QOC) and regularised time (TOC) objective functions.
		For the presented example, the initial state is ($x=-5, z=0, v_x=2, v_z=2, \theta=0$) and the target state is $\mathbf{0}$ (green dot).
		(a) State trajectory plotted for the position variables $(x,z)$ with the pitch angle indicated.
		(b) Corresponding optimal control profiles.
		We note that with respect to the thrust profile (top right), QOC is continuous and TOC is bang-bang.
	}
	\label{fig:control_profiles}
\end{figure}

When solving the time optimal control problem,
the resulting trajectories raised a few issues.
The first was the occurrence of chattering in the control profiles.
The second was the very aggressive nature of the time optimal manoeuvres including  mid-flight flips, which we wanted to avoid as they limit the use of our results on real quadcopters where such manoeuvres would be considered unnecessarily dangerous.
By penalising the time objective with a weighted cost functional
quadratic in $\omega$, we were able to 
eliminate chattering in the profiles of both control inputs.
The weighting factor $\alpha$ was tuned such that
chattering was removed across a
range of choices for the initial and final state.
The regularised time objective function is:
\begin{equation}
    J = T + \alpha \int_0^T \omega(t)^2 \, \text{d}t, \quad \text{where} \; \alpha = 0.1
\end{equation}
Clearly this has the effect of making the resulting 
solutions sub-optimal with respect to the original objective function. The value of $\alpha$ was thus chosen 
such that the optimal time was only marginally increased.
Convergence in the time-optimal case was also improved using the following continuation procedure: 
we first solve for the original problem and then
use the solution as an initial guess for the regularised problem.
Preventing flips means keeping the orientation of the quadcopter close to upright.
This was achieved by adding the following path constraint: $| \theta | \leq \theta^{max}$.
An example trajectory for both objective functions is shown in Fig. \ref{fig:control_profiles}.
A summary of all numerical parameters and their values\footnote{Values used are based on the Parrot Bebop drone} is stated in Table \ref{table:2}.

\begin{table}[!htbp]
\centering
\caption{Numerical parameters of the quadcopter model}
\begin{tabular}{l@{\qquad}l@{\qquad}l}
  \toprule
  Parameter & Value & Description \\
  \midrule
  $m$ & $\SI[group-digits=false]{0.38905}{\kilogram}$ 
    & Mass of the quadcopter \\
  $g$ & $\SI[group-digits=false]{9.81}{\metre\per\second}$ 
    & Acceleration due to Earth's gravity \\
  $\beta$ & $\SI[group-digits=false]{0.5}{}$ 
    & Drag coefficient \\
  $\theta^{max}$ & $\SI[group-digits=false]{\pi / 4}{\radian}$ 
    & Maximum pitch angle \\
  $F_T^{max}$ & $\SI[group-digits=false]{9.1}{\newton}$ 
    & Maximum thrust \\
  $\omega^{max}$ & $\SI[group-digits=false]{35}{\radian\per\second}$ 
    & Maximum pitch rate \\
  \bottomrule
\end{tabular}
\label{table:2}
\end{table}

%% file: 03_policy.tex
For each control task (either quadratic or time objective functions),
we seek to learn a close approximation to the optimal state-feedback map
that navigates the quadcopter from arbitrary initial states, 
within a specified region of the state space,
to some fixed target state.
Optimal trajectories are generated using the method described in \ref{sec:quadcopter}.
These can be considered demonstrations of the optimal policy $\boldsymbol{\pi^{*}}$
from which a regressor is trained to approximate the mapping from state to optimal control.
We first define our state space 
as the compact set $\mathcal{X} \subset \rm I\!R^{n_x}$
where closed intervals are specified for each dimension (see Table \ref{table:3}).
This constrains the optimal trajectories to belong to this region.
The intervals have changed with respect to our previous work \cite{sanchez2018real}.
Since we consider the problem of quadcopter manoeuvres rather than landing, 
the intervals now contain the target state and the endpoints are 
equidistant from the target.
Then $M$ states are drawn uniformly,
$\mathbf{x}_o^{(i)} \sim U(\mathcal{X}) \; \mathrm{for} \; i = 1, ..., M$,
each corresponding to an initial state 
of a trajectory optimisation problem to be solved
(the final state is kept constant).
$M$ must be chosen suitably large as to ensure a dense coverage of the state space.
\begin{table}[!htbp]
\centering
\caption{State space intervals}
\begin{tabular}{c@{\qquad}l}
  \toprule
  State variable & Interval \\
  \midrule
  $x$  & $\interval{-10}{10} \, \si{\metre}$ \\
  $z$  & $\interval{-10}{10} \, \si{\metre}$ \\
  $v_x$ & $\interval{-5}{5} \, \si{\metre\per\second}$ \\
  $v_z$ & $\interval{-5}{5} \, \si{\metre\per\second}$ \\
  $\theta$ & $\interval{-\frac{\pi}{4}}{\frac{\pi}{4}} \, \si{\radian}$ \\
  \bottomrule
\end{tabular}
\label{table:3}
\end{table}

For each trajectory optimisation problem, the solver outputs a sequence of state-control pairs of the form:
\begin{equation}
    \uptau_i = {(\mathbf{x}_j^{(i)}, \mathbf{u}_j^{(i)})}_{j=1}^J \quad \text{where} \; \mathbf{x}_1^{(i)} = \mathbf{x}_o^{(i)}, \; \mathbf{x}_J^{(i)} = \mathbf{x}_f 
    \qquad i = 1, ..., M
\end{equation}
where $J$ is the number of points on a uniformly spaced grid.
The level of discretisation is determined by the number of nodes $K$ chosen in the direct transcription method.
Since the Hermite-Simpson transcription method was used, the midpoints of the nodes are also evaluated thus the number of grid points $J$ is $2K - 1$.
A summary of the parameters used for the large-scale trajectory generation and the values chosen
is provided in Table \ref{table:4}; this is the same for both objective functions.
\added[id=r14]{%
As stated in Section \ref{sec:quadcopter}, the optimal trajectory could be found with the solver simply presented with an arbitrary initial guess for most initial conditions.
This means each trajectory optimisation problem could be solved independently allowing for parallelization of trajectory generation thus speeding up the process and resulting in a considerably larger dataset than previously considered.
We also note that an unintended benefit of this approach is the distribution of initial conditions has more uniform coverage of the state space.
This is a consequence of directly sampling the initial conditions rather than the random walk approach adopted in [6].}
For each control task, it took approximately 1 hour to generate the corresponding library of trajectories when distributed over 40 CPUs.
\begin{table}[!htbp]
\centering
\caption{Dataset generation parameters}
\begin{tabular}{l@{\qquad}l@{\qquad}l}
  \toprule
  Parameter & Value & Description \\
  \midrule
  $x_f$ & $\mathbf{0}$    & Final state \\
  $M$   & $\SI{200000}{}$ & Number of trajectories \\
  $K$   & $30$            & Number of nodes \\
  \bottomrule
\end{tabular}
\label{table:4}
\end{table}

We then construct the datasets by converting each library of trajectories into a collection of state-control pairs,
${\{(\mathbf{x}_i, \mathbf{u}_i)\}}_{i=1}^{N}$ with $N = M \cdot J$,
where trajectory information and ordering is discarded.
Since the trajectories have the same final state, the distribution of inputs within the datasets is skewed towards the target state.
Despite this, we find that the lack of uniformity does not pose a problem 
when fitting a regressor to approximate the map from states to controls.
Once trained, the regressor \textit{is} a deterministic policy $\boldsymbol{\pi} : \mathcal{X} \to \mathcal{U}$
where $\mathcal{X}$ is the state space defined previously and 
$\mathcal{U}$ is the compact set corresponding to the constraints in Table \ref{table:2}.



%% file: 04_dnn.tex
We approach this in the standard machine learning way.
Firstly each dataset is partitioned into a training set (90\%) and a held-out test set (10\%) such that state-control pairs in each set come from distinct trajectories.
The datasets are also preprocessed by way of scaling features of both the inputs and targets to have zero mean and unit variance.

There are many machine learning algorithms appropriate for this task.
We choose to train neural networks as regressors.
This is motivated by their high degree of flexibility (e.g. number of hidden layers, number of units per layer etc.) allowing for a comparison based on the amount of parameterisation.
Furthermore, given the size of the dataset generated, we can use specialised libraries for neural networks that can take advantage of GPUs for faster training.
In our case, we perform neural network training in Keras executed on a NVIDIA GeForce GTX 1080 Ti GPU.
We restrict our attention to feedforward, fully-connected networks with equal number of units in the hidden layers and a squared error loss function.
In contrast to previous work \cite{sanchez2018real}, we consider the problem of jointly learning both targets (controls) in a single regressor.
\added[id=r44]{%
It is worth noting that since the thrust control for time-optimal control has a bang-bang structure, the learning problem for this particular target could be structured as a classification task.
We decided against pursuing this as we seek to demonstrate a policy learning pipeline applicable to the broadest set of control problems.}

From previous work, we take the best performing activation functions, that is the non-saturating rectified linear non-linearity (ReLU) \cite{nair2010rectified}, \added[id=r2]{with analytic form $f(x) = \max(0,x)$}, for the hidden layers and hyperbolic tangent for the output layer.
Since the hyperbolic tangent is bounded, the targets in the dataset are further scaled to lie within the range [-1,1].
Weights are initialised using Xavier's uniform method \cite{glorot2010understanding}.
We use the Adam optimiser, a variant of gradient descent optimisation, with its default values \cite{kingma2014adam}.
This was settled on after a comparison with plain stochastic gradient descent with momentum, a widely used non-adaptive optimiser.

The minibatch size was fixed to 512; this was chosen after a number of preliminary trials.
We found that reducing the minibatch size to very small values consistently resulted in a substantial drop in the performance.
This is in contrast to accepted deep learning practice that smaller minibatch sizes result in better generalisation \cite{lecun2012efficient}.
We partition each training set giving a validation set (10\% held out) that is used to track performance during training.
This is necessary for early stopping for which training ends if no improvement in the validation loss is observed for more than 5 epochs.
The state of the network is saved after every epoch and the final parameter values selected at the end of a training run are those that gave the smallest validation loss.
We also use a learning rate schedule in which the learning rate drops by a factor of 2 if there is no improvement in the validation loss for 3 epochs.

\begin{figure}[!htbp]
\centering
\includegraphics[width=\linewidth]{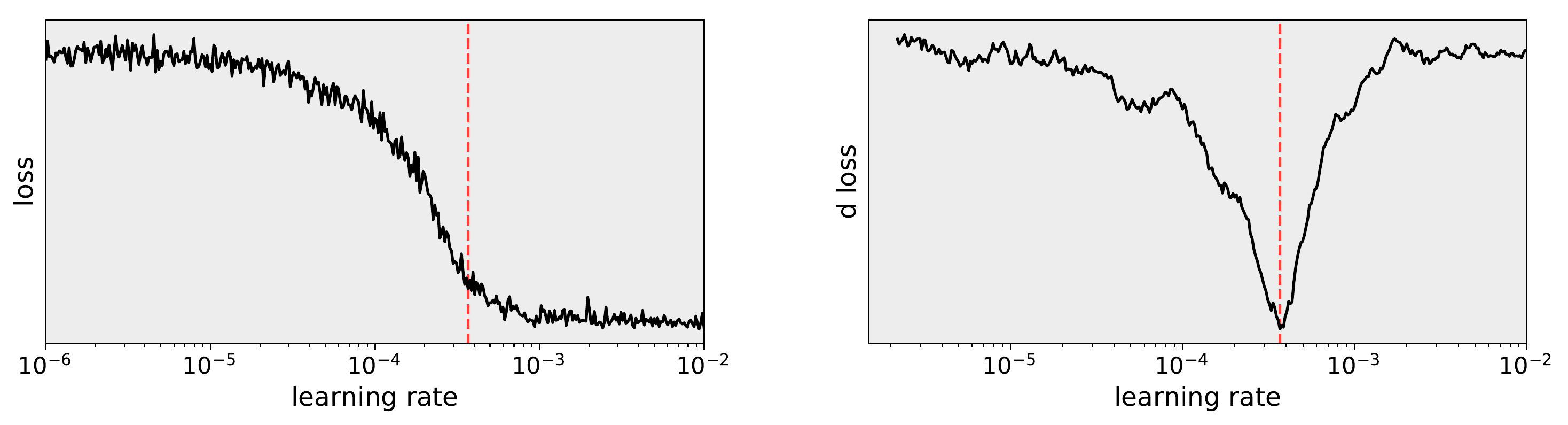}
\caption{Learning rate finder method from \cite{smith2017cyclical}. Learning rate is increased on a logarithmic scale during training of a single epoch starting from a small value. Optimal learning rate is the one that gives the largest decrease in the loss. \textbf{Left}: Initially with small learning rates, the loss improves slowly. As the learning rate increases, the loss improves faster until it explodes (not shown) when the learning rate becomes very large. Fastest decrease in the loss is when the slope of the loss curve is the most negative (dotted red line). \textbf{Right}: Plot of the change in loss per iteration with a moving average filter applied. The optimal learning rate is the minimum of this curve (dotted red line).}
\label{fig:lr_finder}
\end{figure}
	
Despite using an adaptive optimiser, the initial learning rate still needs to be appropriately set.
The learning rate is one of the most important hyperparameters and the success of a given training run is largely determined by it \cite{bengio2012practical}.
Given that we plan to experiment with many network architectures, performing a grid (or better random) search on the learning rate would be too time consuming.
We instead use a simple but relatively unknown method from \cite{smith2017cyclical} to determine a suitable initial learning rate.
This works by training for a single epoch, with the learning rate initially set to a very small value (e.g. $10^{-6}$), whilst increasing the learning rate at each iteration (i.e. minibatch update).
The particular value of the learning rate that gives rise to the largest decrease in the loss is the optimal learning rate (ignoring stochasticity due to the order and batching of the training examples) - see Fig. \ref{fig:lr_finder}.
Since the learning rate is decayed during training, we round the optimal rate up to the nearest power of 10 and set this to be the initial learning rate in the full training run.
For the majority of networks, the initial learning rate was set to $10^{-3}$; only for the deeper and wider architectures did the method indicate a smaller initial learning rate of $10^{-4}$.

%% file: 05_experiments.tex
\begin{table}[!htbp]
\centering
\renewcommand{\arraystretch}{0.7}
\caption{Mean absolute error on the test set evaluated for each dataset: Quadratic optimal control (QOC) and Time optimal control (TOC). Neural networks are compared by their architecture where ``units'' and ``layers'' in the 1st column heading refer to the number of units per hidden layer and the number of hidden layers respectively. The ``normalised'' column gives the error on the dataset after preprocessing and averaged for both targets. The columns ``$u_1$'' and ``$u_2$'' give the error on the datasets before preprocessing evaluated for each target separately. For each number of units considered, the network architecture with the best performance is highlighted for both datasets.}
\begin{tabular}{@{}llllclll@{}}
\toprule
\multirow{2}{3.4cm}{\centering \textbf{Network architecture} \textless units\textgreater -\textless layers\textgreater} 
                                & \multicolumn{3}{c}{\textbf{QOC error}} & \phantom{a} & \multicolumn{3}{c}{\textbf{TOC error}} \\
                                        \cmidrule{2-4} \cmidrule{6-8}
                                       & \multirow{1}{1.2cm}{normalised} 
                                       & \multirow{1}{1.2cm}{$u_1$}     
                                       & \multirow{1}{0.9cm}{$u_2$}     &
                                       & \multirow{1}{1.2cm}{normalised} 
                                       & \multirow{1}{1.2cm}{$u_1$}     
                                       & \multirow{1}{0.9cm}{$u_2$}     \\
                                        \midrule
50-1                               & 0.0293     & 0.2132 & 0.3931 &  & 0.0506     & 0.3737 & 0.3957 \\
50-2                               & 0.0127     & 0.0903 & 0.1839 &  & 0.0282     & 0.2096 & 0.2120 \\
50-3                               & 0.0094     & 0.0667 & 0.1364 &  & 0.0224     & 0.1711 & 0.1480 \\
50-4                               & 0.0096     & 0.0683 & 0.1394 &  & 0.0203     & 0.1509 & 0.1541 \\
50-5                               & 0.0077     & 0.0554 & 0.1055 &  & 0.0197     & 0.1463 & 0.1509 \\
50-6                               & 0.0075     & 0.0543 & 0.1008 &  & 0.0187     & 0.1377 & 0.1464 \\
50-7                               & 0.0073     & 0.0531 & 0.0995 &  & 0.0188     & 0.1382 & 0.1476 \\
50-8                               & 0.0072     & 0.0527 & 0.0978 &  & 0.0186     & 0.1368 & 0.1474 \\
50-9                               & 0.0072     & 0.0526 & 0.0966 &  & 0.0186     & 0.1372 & 0.1438 \\
50-10\textsuperscript{QOC,TOC}                              & \cellcolor{lightgray}0.0072     & \cellcolor{lightgray}0.0523 & \cellcolor{lightgray}0.0968 &  & \cellcolor{lightgray}0.0185     & \cellcolor{lightgray}0.1372 & \cellcolor{lightgray}0.1422 \\
\midrule
100-1                               & 0.0223     & 0.1586 & 0.3260 &  & 0.0455     & 0.3342 & 0.3606 \\
100-2                               & 0.0087     & 0.0635 & 0.1141 &  & 0.0252     & 0.1963 & 0.1498 \\
100-3                               & 0.0070     & 0.0516 & 0.0909 &  & 0.0185     & 0.1439 & 0.1105 \\
100-4                               & 0.0067     & 0.0496 & 0.0858 &  & 0.0181     & 0.1379 & 0.1197 \\
100-5                               & 0.0066     & 0.0487 & 0.0835 &  & 0.0175     & 0.1337 & 0.1162 \\
100-6                               & 0.0066     & 0.0487 & 0.0821 &  & 0.0175     & 0.1334 & 0.1172 \\
100-7                               & 0.0066     & 0.0485 & 0.0818 &  & 0.0174     & 0.1325 & 0.1152 \\
100-8                               & 0.0065     & 0.0482 & 0.0817 &  & 0.0174     & 0.1322 & 0.1184 \\
100-9\textsuperscript{TOC}                               & 0.0065     & 0.0481 & 0.0825 &  & \cellcolor{lightgray}0.0174     & \cellcolor{lightgray}0.1320 & \cellcolor{lightgray}0.1174 \\
100-10\textsuperscript{QOC}                              & \cellcolor{lightgray}0.0065     & \cellcolor{lightgray}0.0481 & \cellcolor{lightgray}0.0818 &  & 0.0174     & 0.1326 & 0.1171 \\
\midrule
200-1                               & 0.0176     & 0.1276 & 0.2385 &  & 0.0406     & 0.3043 & 0.2952 \\
200-2                               & 0.0077     & 0.0569 & 0.0961 &  & 0.0220     & 0.1775 & 0.1040 \\
200-3                               & 0.0064     & 0.0475 & 0.0778 &  & 0.0175     & 0.1377 & 0.0980 \\
200-4                               & 0.0081     & 0.0597 & 0.1021 &  & 0.0169     & 0.1318 & 0.1007 \\
200-5                               & 0.0063     & 0.0467 & 0.0761 &  & 0.0168     & 0.1297 & 0.1055 \\
200-6\textsuperscript{TOC}                               & 0.0062     & 0.0462 & 0.0754 &  & \cellcolor{lightgray}0.0166     & \cellcolor{lightgray}0.1292 & \cellcolor{lightgray}0.0976 \\
200-7                               & 0.0062     & 0.0459 & 0.0744 &  & 0.0166     & 0.1290 & 0.1018 \\
200-8                               & 0.0062     & 0.0461 & 0.0749 &  & 0.0167     & 0.1294 & 0.1008 \\
200-9                               & 0.0062     & 0.0460 & 0.0745 &  & 0.0169     & 0.1312 & 0.1043 \\
200-10\textsuperscript{QOC}                              & \cellcolor{lightgray}0.0061     & \cellcolor{lightgray}0.0458 & \cellcolor{lightgray}0.0742 &  & 0.0167     & 0.1298 & 0.1021 \\
\midrule
500-1                               & 0.0137     & 0.1004 & 0.1759 &  & 0.0353     & 0.2749 & 0.2119 \\
500-2                               & 0.0065     & 0.0488 & 0.0793 &  & 0.0195     & 0.1563 & 0.0952 \\
500-3                               & 0.0062     & 0.0462 & 0.0745 &  & 0.0169     & 0.1340 & 0.0908 \\
500-4                               & 0.0061     & 0.0456 & 0.0734 &  & 0.0165     & 0.1295 & 0.0945 \\
500-5\textsuperscript{TOC}                               & 0.0060     & 0.0451 & 0.0724 &  & \cellcolor{lightgray}0.0161     & \cellcolor{lightgray}0.1268 & \cellcolor{lightgray}0.0889 \\
500-6                               & 0.0067     & 0.0492 & 0.0866 &  & 0.0163     & 0.1283 & 0.0906 \\
500-7                               & 0.0060     & 0.0447 & 0.0713 &  & 0.0172     & 0.1333 & 0.1063 \\
500-8                               & 0.0059     & 0.0444 & 0.0691 &  & 0.0163     & 0.1276 & 0.0942 \\
500-9                               & 0.0059     & 0.0443 & 0.0692 &  & 0.0220     & 0.1653 & 0.1582 \\
500-10\textsuperscript{QOC}                              & \cellcolor{lightgray}0.0059     & \cellcolor{lightgray}0.0442 & \cellcolor{lightgray}0.0692 &  & 0.0164     & 0.1285 & 0.0945 \\
\midrule
1000-1                               & 0.0142     & 0.1058 & 0.1722 &  & 0.0310     & 0.2478 & 0.1546 \\
1000-2                               & 0.0064     & 0.0476 & 0.0759 &  & 0.0185     & 0.1485 & 0.0916 \\
1000-3                               & 0.0061     & 0.0455 & 0.0720 &  & 0.0166     & 0.1311 & 0.0888 \\
1000-4                               & 0.0060     & 0.0448 & 0.0709 &  & 0.0160     & 0.1267 & 0.0864 \\
1000-5                               & 0.0059     & 0.0443 & 0.0696 &  & 0.0161     & 0.1274 & 0.0863 \\
1000-6\textsuperscript{TOC}                               & 0.0059     & 0.0441 & 0.0689 &  & \cellcolor{lightgray}0.0160     & \cellcolor{lightgray}0.1269 & \cellcolor{lightgray}0.0853 \\
1000-7                               & 0.0058     & 0.0438 & 0.0679 &  & 0.0161     & 0.1271 & 0.0872 \\
1000-8                               & 0.0058     & 0.0439 & 0.0682 &  & 0.0161     & 0.1266 & 0.0904 \\
1000-9\textsuperscript{QOC}                               & \cellcolor{lightgray}0.0058     & \cellcolor{lightgray}0.0435 & \cellcolor{lightgray}0.0672 &  & 0.0162     & 0.1268 & 0.0916 \\
1000-10                              & 0.0058     & 0.0439 & 0.0680 &  & 0.0196     & 0.1495 & 0.1303 \\
\bottomrule
\end{tabular}
\label{table:5}
\end{table}

We train a large number of networks with different architectures all following the same training procedure described in Section \ref{sec:dnn}.
We consider networks with the hyperparameters: no. of units per hidden layer: \{50,100,200,500,1000\}; no. of layers: [1..10].
The purpose of this is to investigate limits on the maximum attainable performance of the datasets.
We evaluate the networks using the mean absolute error (MAE) metric, shown in Table \ref{table:5}, and from now on `error' should be read as meaning MAE.

We should note that the errors are not definitive for each architecture.
This is because we did not repeat training with different random seeds and the other hyperparameters (e.g. minibatch size) were not fine-tuned.
\added[id=r16]{%
Despite this, the random seed for each network was initialised independently and so the results are demonstrative of how changes in the network architecture affect performance.}

Consistent with previous work \cite{sanchez2018real}, when comparing the normalised errors between the two datasets, we observe the error for quadratic optimal control (QOC) to be consistently smaller than that of time optimal control (TOC).
This is not surprisingly as the controls (particularly the thrust) is continuous with saturated regions for QOC and bang-bang for TOC.

For a given number of units, we observe that, as the number of layers increases, the error initially decreases as expected.
However, after approximately 5 hidden layers, there is little further improvement - the error saturates or in a few cases increases.
We also observe a difference in behaviour here between the datasets, that is for QOC, the best architecture consistently appears to be the one with 10 layers albeit only small improvement compared with 5 layers or more.
However, for TOC, we observe that with 200 units or more, the best performing network is the one with 5 or 6 hidden layers.
In all, this suggests that stacking more layers is ineffective after a certain point and an increase in the number of units is necessary to gain an improvement in performance.

\sidecaptionvpos{figure}{c}

\begin{SCfigure}
\centering
\includegraphics[width=0.6\linewidth]{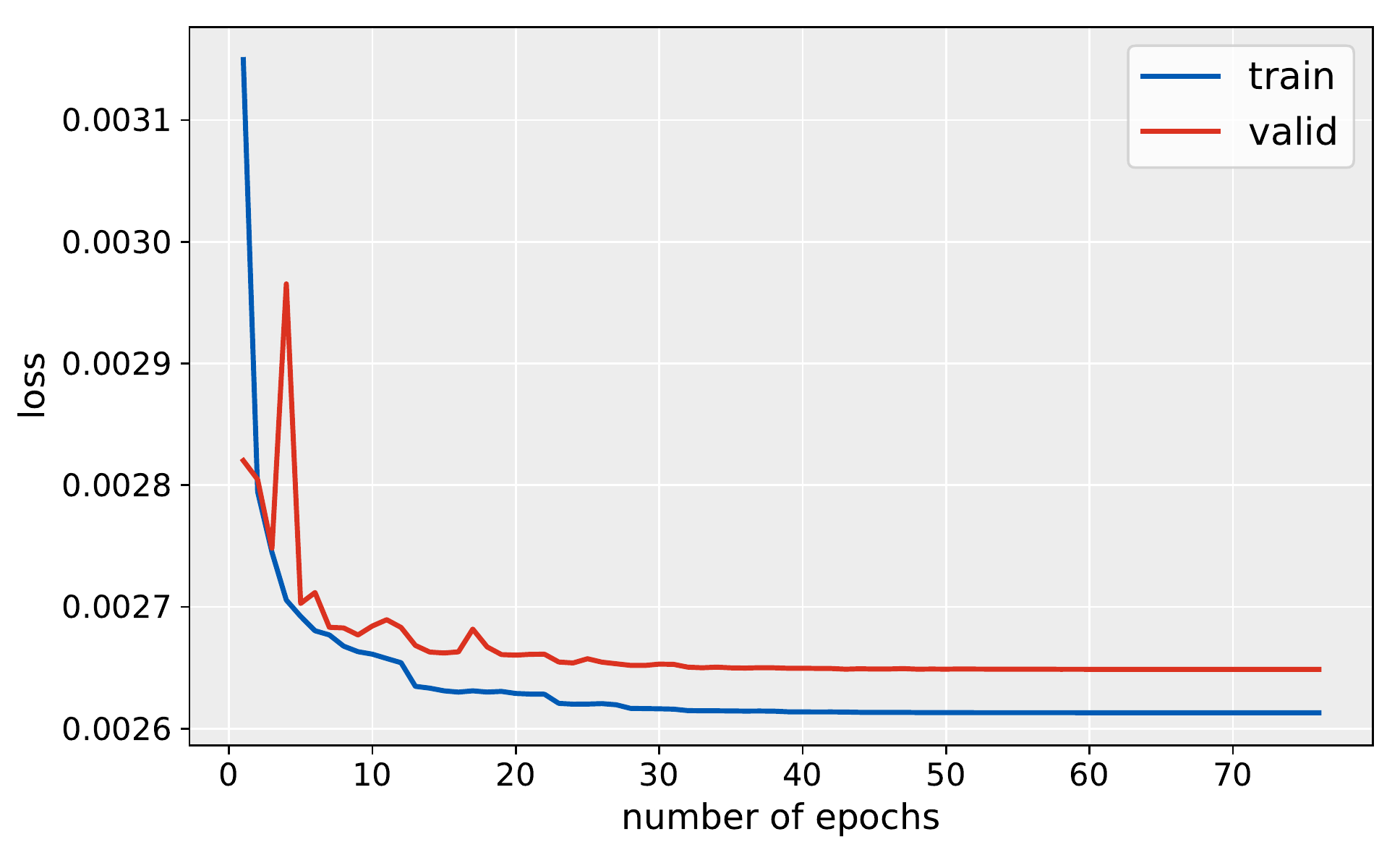}
\caption{Training and validation set loss, as a function of the number of training epochs, for the neural network with 1000 units per hidden layer and 10 hidden layers evaluated on the quadratic optimal control dataset.}
\label{fig:learning_curve}
\end{SCfigure}

By looking at the final values of the training and validation loss across all networks, we observe that overfitting is very minor.
For instance the highest parameterised network trained on QOC overfit by under 2\%.
This learning curve is shown in Fig. \ref{fig:learning_curve}.
We can see that from approximately 30 epochs the training loss plateaus.
Despite training continuing for 40 more epochs, the increase in the validation loss is not considerable.
This indicates that we are still operating in an underfitting regime suggesting further improvement is possible.
It is likely the reason for the underfitting is the large training set size.
Considering the highest parameterised network, this has fewer parameters than the number of training set datapoints (9,017,002 vs 9,440,000).
Thus, the training set size had a regularising effect eliminating the need for techniques such as DropOut \cite{srivastava2014dropout} or weight decay.

We found that adding more layers led to a higher training loss.
This \textit{degradation} problem is a common occurrence in deep learning.
Methods have been developed in recent years to alleviate this such as batch normalisation \cite{ioffe2015batch} and residual connections \cite{he2016deep}.
We decided against pursuing this due to the saturation observed with the depths already considered.
Although it is likely we would have been able to further reduce the error with more units per layer, we also observed saturation in the test error - see the shaded rows in Table \ref{table:5}.

%% file: 06_policy_eval.tex
For a given dataset, the different neural networks correspond to different instances of the same control policy.
As shown in Table \ref{table:5}, the neural networks vary in their accuracy at reproducing the state-to-control mapping of the optimal trajectories.
Even the highest parameterised neural network does not reproduce the mapping exactly exhibiting a minimum error.
Therefore, the resulting policies can only be considered to be approximations to the optimal state-feedback.
Despite this, the difference between prediction and ground truth is small when compared to the range of control values that can be attained.

The policies must also be evaluated on their ability to perform the control task.
By numerical integration of the system dynamics with the policy substituted for the control inputs we can compute the state trajectory,
\begin{equation}
    \mathbf{x}(t) = \mathbf{x}_o + \int_0^t \mathbf{f}(\mathbf{x}(\uptau), \boldsymbol{\pi}(\mathbf{x}(\uptau))) \, \text{d}\uptau,
\end{equation}
starting from any initial state $\mathbf{x}(0) = \mathbf{x}_o$.
This can be seen as a closed-loop (feedback) control strategy.
Clearly, the control trajectory can also be computed by evaluating the policy at predefined intervals along the state trajectory.

We repeat the analysis undertaken in \cite{sanchez2018real} and find that all the policies (including the lowest parameterised neural networks) eventually reach the target state with small error.
This is the case regardless of the initial state, provided it is contained within the state space.
Also the resulting trajectories closely resemble the optimal trajectory for the equivalent manoeuvre.
After the target state is reached, the system stabilises at a state close to the target in terms of $x$ and $z$.
The $v_x$, $v_z$ and $\theta$ components must be zero.
This \textit{hovering} phase, reported in \cite{sanchez2018real}, is achieved when the thrust predicted by the neural network cancels $m \cdot g$ and the pitch rate takes a value close to $0$, as determined by the quadcopter dynamics.

We now further investigate the trained policies by looking at the performance up to when it reaches the target (the \textit{manoeuvre} phase) and the behaviour during the hovering phase.

\subsection{Trajectory optimality}
We shall investigate whether there is any relationship between the accuracy of a policy's state-feedback approximation and its performance on the control task.
For this we look at relative optimality, one of the evaluation metrics used in previous work \cite{sanchez2018real}.
This involves evaluating the objective function $J$ on a state-control trajectory resulting from some policy and then computing the relative error of this value with respect to that of the optimal trajectory.
This is then repeated for multiple manoeuvres giving an average relative optimality.

We have improved the algorithm for determining relative optimality.
Previously, a fixed integration time was used regardless of distance from the target state.
The trajectory was then truncated such that the final state is the one closest to the target. 
This raised a few issues.
The first is that, in the distance metric, the state variables were scaled to overcome the problem of unit heterogeneity.
This made the trajectory time sensitive to the choice of scale factors.
More concerning was that if the system stabilises near the target then the closest state might be reached during the hovering phase.
This could skew the results as the objective value of the optimal trajectory is restricted to the manoeuvre.
To overcome these issues, we use, for the integration time, the final time of the corresponding optimal trajectory.
Full details are outlined in algorithm \ref{alg:optimality}.

\noindent
\begin{minipage}{.75\linewidth}
\begin{algorithm}[H]
\setstretch{1.1}
\caption{Policy trajectory optimality}
\label{alg:optimality}
\begin{algorithmic}[1]
\STATE Given initial state $\mathbf{x}_o$ and final time $t_f$ of trajectory from test set
\STATE Compute trajectory arising from policy with initial state $\mathbf{x}_o$ and integration time $t_f$
\STATE Extract final state $\mathbf{x}_f$ from policy trajectory
\STATE Solve for optimal trajectory with initial state $\mathbf{x}_o$ and final state $\mathbf{x}_f$
\STATE Evaluate objective function on policy trajectory and optimal trajectory: $j^{\pi}$, $j^{*}$
\STATE Compute relative error: $(j^{\pi} - j^{*}) / j^{*}$
\end{algorithmic}
\end{algorithm}
\end{minipage}
\vspace{4mm}


\begin{table}[!htbp]
\centering
\caption{Relative error of each objective function, quadratic optimal control (QOC) and time optimal control (TOC), evaluated on the policy trajectory with respect to the objective value of the optimal trajectory averaged over trajectories from the test set. This is shown for policies represented by neural networks with 100 units per hidden layer accompanied by the corresponding test set error.}
\begin{tabular}{cccccc}
  \toprule
  \multirow{2}{3.5cm}{\centering \textbf{Network architecture} \textless units\textgreater -\textless layers\textgreater} & \multicolumn{2}{c}{\textbf{QOC}} & \phantom{a} & \multicolumn{2}{c}{\textbf{TOC}} \\
  \cmidrule{2-3} \cmidrule{5-6}
  & MAE & Optimality (\%) & & MAE & Optimality (\%) \\
  \midrule
  100-1 & 0.0223 & 1.674 & & 0.0455 & 2.475 \\
  100-2 & 0.0087 & 0.266 & & 0.0252 & 0.900 \\
  100-3 & 0.0070 & 0.199 & & 0.0185 & 0.605 \\
  100-4 & 0.0067 & 0.206 & & 0.0181 & 0.793 \\
  100-5 & 0.0066 & 0.238 & & 0.0175 & 0.664 \\
  100-6 & 0.0066 & 0.241 & & 0.0175 & 0.759 \\
  100-7 & 0.0066 & 0.299 & & 0.0174 & 0.648 \\
  100-8 & 0.0065 & 0.220 & & 0.0174 & 0.707 \\
  100-9 & 0.0065 & 0.239 & & 0.0174 & 0.740 \\
  100-10 & 0.0065 & 0.207 & & 0.0174 & 0.669 \\
  \bottomrule
\end{tabular}
\label{table:optimality}
\end{table}

In general, we observe that all the policies, even those represented by shallow neural networks,
are close-to-optimal.
Furthermore, the quadratic-optimal policies have a smaller relative optimality error
than the time-optimal policies.
This is consistent with the state-feedback approximation performance between the two datasets observed previously.

The error between the optimal cost and the policy cost appears to initially reduce as the corresponding policy's state-feedback approximation error decreases (see Table \ref{table:optimality}).
However, the relationship does not continue for neural networks with greater depths despite their state-feedback approximation error continuing to decrease.
We should emphasise that the improvement in state-feedback approximation is not very considerable after a network depth of 3.
The results indicate that despite similar state-feedback approximations, when such errors are propagated during integration of the system, they result in varying values of the trajectory cost.
This shows that finding the regressor with the lowest state-feedback approximation error is not necessary as small improvements are not reflected in the policy optimality. 

We should highlight that the errors are rather small, with the time optimal case resulting in a time loss of a few milliseconds at most in the best networks. This confirms the extreme precision achievable by the pursued approach, at least on the low dimensional dynamics here considered.

\subsection{Asymptotic behaviour}
We find that the system stabilises to the same point regardless of the initial state. 
The equilibrium point $\mathbf{x_e} $ for each policy is defined by $\mathbf{f}(\mathbf{x_e}, \boldsymbol{\pi}(\mathbf{x_e})) = \mathbf 0$.

A consequence of the dataset generation process is that the target state is by far the majority state in the datasets.
However, the corresponding control for this state across instances is not identical.
This is a consequence of the optimal control solver used which, for the control at the target state, simply gives a smooth continuation of the preceding controls in the trajectory.
In fact, the change in the intervals, particularly the $z$ dimension, from which the initial states of the optimal trajectories were sampled from resulted in a shift in the distribution of control values at $\mathbf{x}_f$.
Although this was not the intention of the change, for both datasets the mean of this distribution turned out to be closer to $[mg, 0]$.
We noticed consistently that the system now stabilised much closer to the target than before.
This suggests that the distance of the equilibrium point from the target is related to the controls predicted by the policy at $\mathbf{x}_f$.
This allows for the possibility of the equilibrium point being influenced at training time.


We determine the equilibrium point for several neural networks and find that, for both control tasks, the network whose policy's equilibrium point is furthest from the target is the one with the lowest accuracy in state-feedback approximation (see `100-1' in Table \ref{table:1}).
A consequence of the quadcopter dynamics is that $\mathbf{x}_e$ for all policies is non-zero only in the $x$ and $z$ dimensions 
and we consider a target state that is zero in all variables.
Since the units are equivalent, any distance metric would be suitable here and hence we use Euclidean distance. 

By considering the definition of stability, for different radii we evaluate the time at which the system first enters into a ball centred at its equilibrium point.
This time is averaged over many different trajectories starting from initial states taken from the test set, giving a \textit{stability time} for each policy and ball radius (Table \ref{table:1}).
To avoid aggregating values with different units, we use the Chebyshev distance as the metric to define the ball.
This constrains all state variables to lie a given distance from $\mathbf{x}_e$.
Given an equilibrium point $\mathbf{x}_e$ and radius $r$, we define the ball as: $B_r(\mathbf{x}_e) = \{\mathbf{x} \in \rm I\!R^{n_x} | \max_i (| x_i - {(\mathbf{x}_e)}_i |) < r \}$.

As expected, we observe the stability time to increase as the ball radius decreases.
We also observe the stability time for the time-optimal policies to be lower for the largest ball radius considered.
For the larger radii, the stability times are similar.
Furthermore, we observe the stability time to be greater for policies where $\mathbf{x}_e$ is further from $\mathbf{x}_f$.
This is clear when comparing the network architecture `100-1' against the others in Table \ref{table:1}.

\begin{table}[!htbp]
\centering
\caption{Average time taken for policies to stabilise at their equilibrium point. This is shown for different radii $r$ and network architectures for both control tasks: quadratic optimal control (QOC) and time optimal control (TOC).}
\begin{tabular}{@{}ccccccc@{}}
  \toprule
  \multirow{2}{3.3cm}{\centering \textbf{Network architecture} \textless units\textgreater -\textless layers\textgreater}
  & \multirow{2}{3.6cm}{\centering \textbf{Distance of equilibrium from target}}
  & \multicolumn{5}{c}{\textbf{Stability time (s)}} \\
  \cmidrule{3-7}
  && $r=10^{-1}$ & $r=10^{-2}$ & $r=10^{-3}$ & $r=10^{-4}$ & $r=10^{-5}$ \\
  \midrule
  \multicolumn{1}{l}{\textbf{QOC}} \\
    100-1              & 0.0118             & 2.728 & 5.228 & 8.910  & 13.102 & 17.200  \\
    100-2              & 0.0023             & 2.455 & 2.513 & 3.592 & 5.089  & 6.486 \\
    100-3              & 0.0025             & 2.453 & 2.509 & 3.094 & 3.634  & 4.421 \\
    100-4              & 0.0057             & 2.453 & 2.540  & 3.561 & 4.502  & 6.034 \\
    100-5              & 0.0025             & 2.452 & 2.518 & 3.090  & 3.779  & 4.481 \\
  \midrule
  \multicolumn{1}{l}{\textbf{TOC}} \\
    100-1              & 0.1058             & 3.021 & 6.003 & 9.252 & 12.454 & 15.398 \\
    100-2              & 0.0302             & 1.987 & 5.052 & 8.316 & 10.876 & 13.370  \\
    100-3              & 0.0138             & 1.957 & 2.340  & 2.905 & 3.447  & 3.991  \\
    100-4              & 0.0180              & 1.963 & 2.426 & 3.317 & 4.104  & 4.920   \\
    100-5              & 0.0192             & 1.958 & 2.530  & 3.290  & 4.035  & 4.788  \\
  \bottomrule
\end{tabular}
\label{table:1}
\end{table}

%% file: 07_softplus.tex
We have shown that neural networks can reproduce the state-to-control mapping of optimal trajectories with low error and that the cost of the policy trajectories is very close to the optimal cost.
For quadratic optimal control, we observe the optimal trajectories to have a control profile that is smooth.
Although having little effect on the state trajectory, the control profile of the policy trajectories exhibit sharp changes in direction.
We should emphasise that the difference here is often indistinguishable, particularly for networks with more than one hidden layer.
The neural networks we have been training thus far are composed of piecewise linear functions (ReLUs).
We hypothesise that the current deficiency is the result of attempting to approximate a smooth function (i.e. the optimal state-feedback for QOC) with a non-smooth function.

We investigate the effect of training neural networks that, for the intermediate units, use a smooth version of ReLU called softplus for the activation function.
This has the analytic form: $f(x) = \log (1 + \exp(x))$.
This keeps useful properties of ReLU such as its unboundedness, necessary for overcoming the vanishing gradient problem.
The performance of such networks trained using the same procedure as in Section \ref{sec:dnn} is comparable to those trained previously (see Table \ref{table:softplus}).
For quadratic optimal control, except for networks with a single hidden layer, we consistently find lower error in the softplus networks.
We can also confirm that the relative optimality of the resulting policies is similar to those reported previously.

\begin{figure}[!htbp]
\centering
\includegraphics[width=\linewidth]{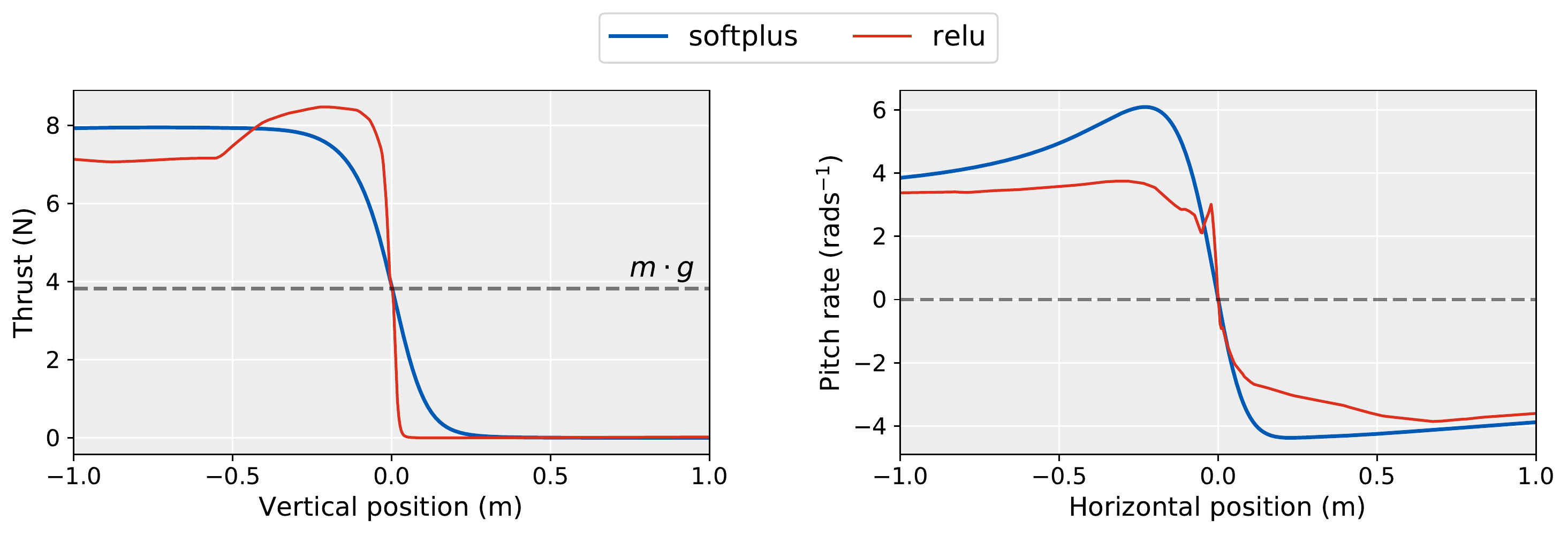}
\caption{Comparison of the control profiles of policies derived from ReLU and softplus neural networks for quadratic optimal control. The controls necessary for the system to be at equilibrium are indicated (grey dashed line).}
\label{fig:softplus}
\end{figure}

In Fig. \ref{fig:softplus}, for two policies corresponding to a ReLU and softplus network, we show how the thrust and pitch rate change when $z$ and $x$ are varied respectively about 0 with the other state variables fixed at the origin.
We observe that, whilst the policies with ReLU and softplus networks behave similarly, the controls for the softplus network have the desireable property of a smooth curve.

\begin{table}[!htbp]
\centering
\caption{Comparison of test set performance between ReLU and softplus neural networks. The metric shown corresponds to the `normalised' error. We highlight networks with the \underline{\textbf{lowest error}}.}
\begin{tabular}{@{}lllcll@{}}
\toprule
\multirow{2}{3.3cm}{\centering \textbf{Network architecture} \textless units\textgreater -\textless layers\textgreater} 
                                & \multicolumn{2}{c}{\textbf{QOC error}} & \phantom{a} & \multicolumn{2}{c}{\textbf{TOC error}} \\
                                        \cmidrule{2-3} \cmidrule{5-6}
                                       & \multirow{1}{1.2cm}{ReLU} 
                                       & \multirow{1}{1.2cm}{softplus}  &   
                                       & \multirow{1}{1.2cm}{ReLU} 
                                       & \multirow{1}{1.2cm}{softplus} \\
                                        \midrule
50-1                & 0.0293    & 0.0317      & & 0.0506         & 0.0558 \\
50-2                & 0.0127    & 0.0113      & & 0.0282         & 0.0302 \\
50-3                & 0.0094    & 0.0079      & & 0.0224         & 0.0281 \\
50-4                & 0.0096    & 0.0075      & & 0.0203         & 0.0218 \\
50-5                & 0.0077    & 0.0079      & & 0.0197         & 0.0200 \\
50-6                & 0.0075    & 0.0065      & & 0.0187         & 0.0184 \\
50-7                & 0.0073    & \underline{\textbf{0.0063}}     & & 0.0188         & 0.0184 \\
50-8                & 0.0072    & 0.0066      & & 0.0186         & 0.0180 \\
50-9                & 0.0072    & 0.0074      & & 0.0186         & 0.0182 \\
50-10               & \underline{\textbf{0.0072}}    & 0.0093      & & \underline{\textbf{0.0185}}         & \underline{\textbf{0.0180}} \\
\midrule
100-1                & 0.0223    & 0.0303      & & 0.0455      & 0.0536   \\
100-2                & 0.0087    & 0.0084      & & 0.0252      & 0.0285   \\
100-3                & 0.0070    & 0.0072      & & 0.0185      & 0.0219   \\
100-4                & 0.0067    & 0.0061      & & 0.0181      & 0.0193   \\
100-5                & 0.0066    & 0.0061      & & 0.0175      & 0.0177   \\
100-6                & 0.0066    & 0.0064      & & 0.0175      & 0.0174   \\
100-7                & 0.0066    & \underline{\textbf{0.0059}}      & & 0.0174      & 0.0174   \\
100-8                & 0.0065    & 0.0060      & & 0.0174      & 0.0170   \\
100-9                & 0.0065    & 0.0060      & & \underline{\textbf{0.0174}}      & 0.0170   \\
100-10               & \underline{\textbf{0.0065}}    & 0.0061      & & 0.0174      & \underline{\textbf{0.0168}}   \\
\midrule
200-1                & 0.0176    & 0.0282      & & 0.0406      & 0.0437   \\
200-2                & 0.0077    & 0.0073      & & 0.0220      & 0.0262   \\
200-3                & 0.0064    & 0.0066      & & 0.0175      & 0.0206   \\
200-4                & 0.0081    & 0.0059      & & 0.0169      & 0.0180   \\
200-5                & 0.0063    & \underline{\textbf{0.0059}}      & & 0.0168      & 0.0174   \\
200-6                & 0.0062    & 0.0059      & & \underline{\textbf{0.0166}}      & \underline{\textbf{0.0168}}   \\
200-7                & 0.0062    & 0.0059      & & 0.0166      & 0.0168   \\
200-8                & 0.0062    & 0.0062      & & 0.0167      & 0.0172   \\
200-9                & 0.0062    & 0.0059      & & 0.0169      & 0.0169   \\
200-10               & \underline{\textbf{0.0061}}    & 0.0061      & & 0.0167      & 0.0240   \\
\bottomrule
\end{tabular}
\label{table:softplus}
\end{table}

%% file: 08_conclusion.tex
We show that the straightforward approach of constructing a near-optimal control policy by
supervised learning on trajectories satisfying Pontryagin's principle of optimality is successful in the low-dimensional problem here considered.
In addition to the low error achieved in the state-feedback approximation of the trajectories,
for a majority of the trained policies,
their optimality, from aggregating the cost from different initial states, 
is just up to a percent worse than that of the optimal.
Furthermore, the policies exhibit various desirable properties 
such as consistent state convergence regardless of initial condition.

We performed large-scale training of many neural network architectures 
on two datasets corresponding to quadratic-optimal and time-optimal
trajectories for a simple quadcopter model in two-dimensions.
In both cases, we were able to saturate the state-feedback approximation error
after a certain amount of parameterisation.
We also conclude that the networks do not necessarily need to be very deep,
finding 5 or 6 hidden layers to be adequate.
For the time-optimal dataset, we found that incorporating even more layers
was detrimental to the final performance.
We did not use any form of regularisation, 
in contrast to standard practice in deep learning,
as we found that the use of large datasets
meant that we were always operating in an underfitting regime.

We only observe a positive relationship
between state-feedback approximation accuracy and
policy optimality for the shallower networks.
This can be understood as initially increasing the number of layers 
gives rise to the largest increase in feedback approximation accuracy.
However, we observe that small gains in the feedback approximation 
does not translate to better optimality.
We show that it is not necessary to find the best feedback approximation.
It is possible to identify a threshold after which
the policy optimality is equivalent
thereby constraining the hyperparameter search on the architecture.

We observe that the system, when controlled by the trained policies,
stabilise at a state unique to each policy.
This allowed us to consider an additional metric, that is the time taken to reach 
this convergent state.
We observe that this convergent state or equilibrium point 
appears to be further from the target state 
for policies with worse state-feedback approximation accuracy.
This also coincides with a longer time to reach the equilibrium point.

The last investigation involved changing the 
activation function of the units in the hidden layers to 
softplus, a smooth approximation to the rectifier.
In addition to giving a similar state-feedback approximation accuracy,
we observe the control profiles of the resulting policies to have a smooth behaviour (for quadratic-optimal control) - 
a property preferable for control systems.
This suggests that best practices in deep learning, such as the use of \replaced[id=r2]{rectifiers}{ReLUs},
are not always transferable to problems in continuous control.

We noted the regularisation effect that training on large datasets had on learning
the optimal state-feedback.
A further avenue of research would be to investigate how this effect varies with different dataset sizes.
From statistical learning theory, the difference between the final training and test set loss values gives an indication of overfitting.
Investigating how this varies with different dataset sizes would help in 
understanding the sample complexity of this method.

In this work, we continued with the same low-dimensional dynamical model used in previous work \cite{sanchez2018real}.
The next logical step in this line of research would be investigating
the effectiveness of this approach for higher dimensional systems.